\documentclass[final,5p,times,twocolumn]{elsarticle}

\makeatletter
\def\ps@pprintTitle{%
	\let\@oddhead\@empty
	\let\@evenhead\@empty
	\let\@oddfoot\@empty
	\let\@evenfoot\@oddfoot
}
\makeatother

\usepackage[switch]{lineno}
\usepackage[hidelinks]{hyperref}

\usepackage{amsmath}
\usepackage{amssymb}

\usepackage{subfigure}
\graphicspath{{figures/}}
\DeclareGraphicsExtensions{.pdf,.jpeg,.png}

\journal{Information Fusion}

\begin{document}

\begin{frontmatter}

\title{EmbraceNet: A robust deep learning architecture for multimodal classification}

\author{Jun-Ho Choi}
\ead{idearibosome@yonsei.ac.kr}

\author{Jong-Seok Lee\corref{cor}}
\ead{jong-seok.lee@yonsei.ac.kr}

\address{School of Integrated Technology, Yonsei University, 85 Songdogwahak-ro, Yeonsu-gu, Incheon, Korea}

\cortext[cor]{Corresponding author}

\begin{abstract}
Classification using multimodal data arises in many machine learning applications.
It is crucial not only to model cross-modal relationship effectively but also to ensure robustness against loss of part of data or modalities.
In this paper, we propose a novel deep learning-based multimodal fusion architecture for classification tasks, which guarantees compatibility with any kind of learning models, deals with cross-modal information carefully, and prevents performance degradation due to partial absence of data.
We employ two datasets for multimodal classification tasks, build models based on our architecture and other state-of-the-art models, and analyze their performance on various situations.
The results show that our architecture outperforms the other multimodal fusion architectures when some parts of data are not available.
\end{abstract}


\end{frontmatter}

\section{Introduction}

Nowadays, deep learning is widely used in many applications \cite{schmidhuber2015deep,krizhevsky2012imagenet,ordonez2016deep,choi2017impact,donahue2015long}.
Within only a decade, various deep learning models have been suggested and investigated, including convolutional neural networks (CNNs) \cite{krizhevsky2012imagenet,choi2017impact,donahue2015long,ji20133d,dosovitskiy2015flownet}, recurrent neural networks \cite{ordonez2016deep,choi2017impact,donahue2015long}, generative adversarial networks \cite{dosovitskiy2015flownet,goodfellow2014generative}, and so on.

One of the most notable triggers that enrich the deep learning approach is the abundance of resources and data.
As the graphics processing units (GPUs) and multiple computing devices have been made easily utilized \cite{schmidhuber2015deep,abadi2016tensorflow}, it became possible to employ large deep learning architectures with huge numbers of layers \cite{he2016deep}.
In addition, an extensive amount of data has accelerated the opportunities of the deep learning, along with the development of big data analysis \cite{chen2014big}.

In conjunction with the expansion of the data availability, more and more sensors have been employed in recent years.
It is feasible to obtain data from sensors in various domains, including microphones, cameras, motion controllers, wearable inertial sensors, and so on \cite{ofli2013berkeley,chen2015utd}.
In addition, thanks to the smartphones equipped with several sensors such as accelerometers and gyroscopes, it became possible to retrieve meaningful multimodal information of users much easier than before \cite{kwon2014unsupervised}.

In real-world situations, however, it is not possible to ensure the availability of all data acquisition devices.
For example, wirelessly connected sensors may be occasionally disconnected and become unable to send any data for the time being \cite{chavarriaga2013opportunity}.
Unfortunately, most deep learning architectures that utilize multimodal data are not properly designed to handle such unexpected situations.
Some deep network models try to fill missing data by repeating previously observed values, using default values, or employing interpolation methods \cite{ordonez2016deep,ngiam2011multimodal}.
However, these methods serve as workarounds that are not the fundamental solution to handle missing data thoroughly.

To tackle this, we propose a novel deep learning architecture called ``EmbraceNet,'' which is designed for multimodal information-based classification tasks in the wild.
The main components of the proposed architecture are \textit{docking} layers, which convert the information of each modality to a representation suitable for integration, and an \textit{embracement} layer, which combines the representations of multiple modalities in a probabilistic manner.
Our architecture provides good compatibility with any network structure, in-depth consideration of correlations between different modalities, and seamless handling of missing data.
Here are the main advantages that the proposed architecture offers.
\begin{itemize}
	\item
	The EmbraceNet model supports high compatibility with existing deep learning architectures.
	Our model takes inputs from any kind of network models, embraces them, and builds a fused representation that can serve as the input of the final decision model.
	This structure enables applying the EmbraceNet architecture to various network models for multimodal classification tasks.
	In addition, any number of modalities can be integrated in our model.
	\item 
	The EmbraceNet model considers cross-modal correlations thoroughly.
	During the training stage, our architecture operates the \textit{embracement} process that probabilistically selects partial information from each modality for combination, which eventually models correlations between different modalities.
	In addition, the learning process is well regularized so that the model effectively avoids overfitting.
	\item 
	The EmbraceNet model ensures robustness against data loss.
	In the trained model, the missing information due to data loss of a modality can be covered by the other modalities.
	Therefore, our model efficiently keeps its performance from both block-wise missing data and even missing the entire data of some modalities.
\end{itemize}

We conduct experiments to demonstrate the robustness of the proposed model on the gas sensor arrays dataset \cite{vergara2013performance} containing data from eight types of gas sensors arranged in nine sensor arrays, and the OPPORTUNITY dataset \cite{chavarriaga2013opportunity} consisting of multi-labeled human activity data obtained from 19 sensors.

The rest of the paper is organized as follows.
We first discuss some related work in Section~\ref{sec:relatedwork}.
Section~\ref{sec:modeldescription} introduces the overall structure of our proposed EmbraceNet model and Section~\ref{sec:benefits} describes several major benefits of the model.
Section~\ref{sec:optimizing} suggests optimization techniques to enhance the overall performance of the EmbraceNet model with in-depth investigation.
We present performance analysis of the model in comparison to other well-known deep learning architectures for the classification tasks, by explaining the experimental setup in Section~\ref{sec:setup} and interpreting the results in Section~\ref{sec:results}.
Finally, we conclude our work in Section~\ref{sec:conclusion}.

\section{Related work}
\label{sec:relatedwork}

Multimodal data have been widely employed in recent decades \cite{poria2017review}.
One of the most prominent multimodal data is videos, which consist of image frames and audio signals \cite{matthews2002extraction,lee2009two,thomee2016yfcc100m}.
In addition, human activity recognition systems usually employ data obtained from multiple sensors, including motion capture system, depth cameras, accelerometers, and microphones \cite{ofli2013berkeley,chen2015utd,chavarriaga2013opportunity}.
There also exist several multimodal datasets in biological \cite{higuera2015self}, chemical \cite{vergara2013performance}, or medical \cite{singh2014fusion} researches.

Many machine learning-based fusion approaches have been proposed to handle multimodal information for classification tasks.
The most widely used conventional techniques are early integration (or data fusion) and late integration (or decision fusion) \cite{lee2008robust,atrey2010multimodal,verma2014multimodal,moon2015perceptual}.
Some studies employed both methods to take their benefits simultaneously \cite{atrey2010multimodal}.

In the early integration method, data of all modalities are concatenated at the initial stage and serve as a single input of the classifier \cite{snoek2005early}.
The data of each modality is usually converted to a feature vector before concatenation via, e.g., principal component analysis \cite{kusuma2011pca} and linear discriminant analysis \cite{ahmad2010multimodal}.
Since the multimodal data serve as a single vector, any classification models that treat unimodal data can be easily adopted.
In addition, the early integration approach considers the cross-modal correlations from the initial stages.
However, it assumes perfect synchronization of different modalities, which may not provide the best performance for tasks requiring flexible synchrony modeling, e.g., audio-visual speech recognition \cite{lee2009two,bengio2004multimodal}.

Unlike the early integration method, the late integration method constructs a separate classifier for each modality, trains the classifiers independently, and draws a final decision by combining outputs of the classifiers \cite{verma2014multimodal}.
Each separate classifier is optimized to the corresponding modality.
There are various ways to make a decision in the late integration method, and it is known that combining with the weighted sum rule usually outperforms the multiplication rule \cite{kittler1998combining}.
The late integration approach does not share any representations across different modalities, which results in ignoring the correlated characteristics among the modalities.

Some researchers developed more sophisticated multimodal integration approaches that regulate the degree of contribution of each modality for classification tasks.
Bharadwaj \textit{et al.} \cite{bharadwaj2015qfuse} proposed a context switching-based person identification system, which prioritizes the multimodal data by measuring their quality and chooses an appropriate classifier.
Goswami \textit{et al.} \cite{goswami2016improving} employed the pool adjacent violators algorithm for combining multiple classifiers, which calibrates the outputs of the classifiers with respect to their confidence values.
Choi and Lee \cite{choi2018confidence} developed an activity recognition model, which controls the amount of information for each modality by measuring its reliability.
These imply that considering significance of each modality data is beneficial to improve the classification performance.

Recently, deep learning has been adopted to model multimodal data.
While the structures to handle data and features are based on deep learning techniques such as convolutional and recurrent neural networks, the conventional early and late integration approaches are still popularly used.
For example, Ord{\'o}{\~n}ez and Roggen employed the early integration method by concatenating all raw sensor channels, which was inputted to a single deep network model consisting of convolutional and recurrent layers \cite{ordonez2016deep}.
Costa \textit{et al.} employed the late integration method by fusing decisions obtained from conventional and deep learning-based approaches for music genre classification \cite{costa2017evaluation}.
On the other hand, some studies introduced intermediate integration methods, where learned representations for different modalities are combined in the middle part of a deep learning network.
For instance, Ngiam \textit{et al.} extracted audio and visual features separately from the given video via dedicated restricted Boltzmann machines (RBMs), concatenated them to construct a shared representation, and fed it to the classifier \cite{ngiam2011multimodal}.
Amer \textit{et al.} extended the intermediate integration-based RBMs to deal with temporal information of the time-series data \cite{amer2018deep}.
Gao \textit{et al.} proposed a method called compact bilinear pooling, where each modality feature is handled by the convolution of count sketches \cite{pham2013fast} at the intermediate stage \cite{gao2016compact}.
These architectures enable not only handling modalities of different domains in a single network, but also considering mid-level information between modalities.

\begin{figure*}[t]
	\centering
	\includegraphics[width=5.2in]{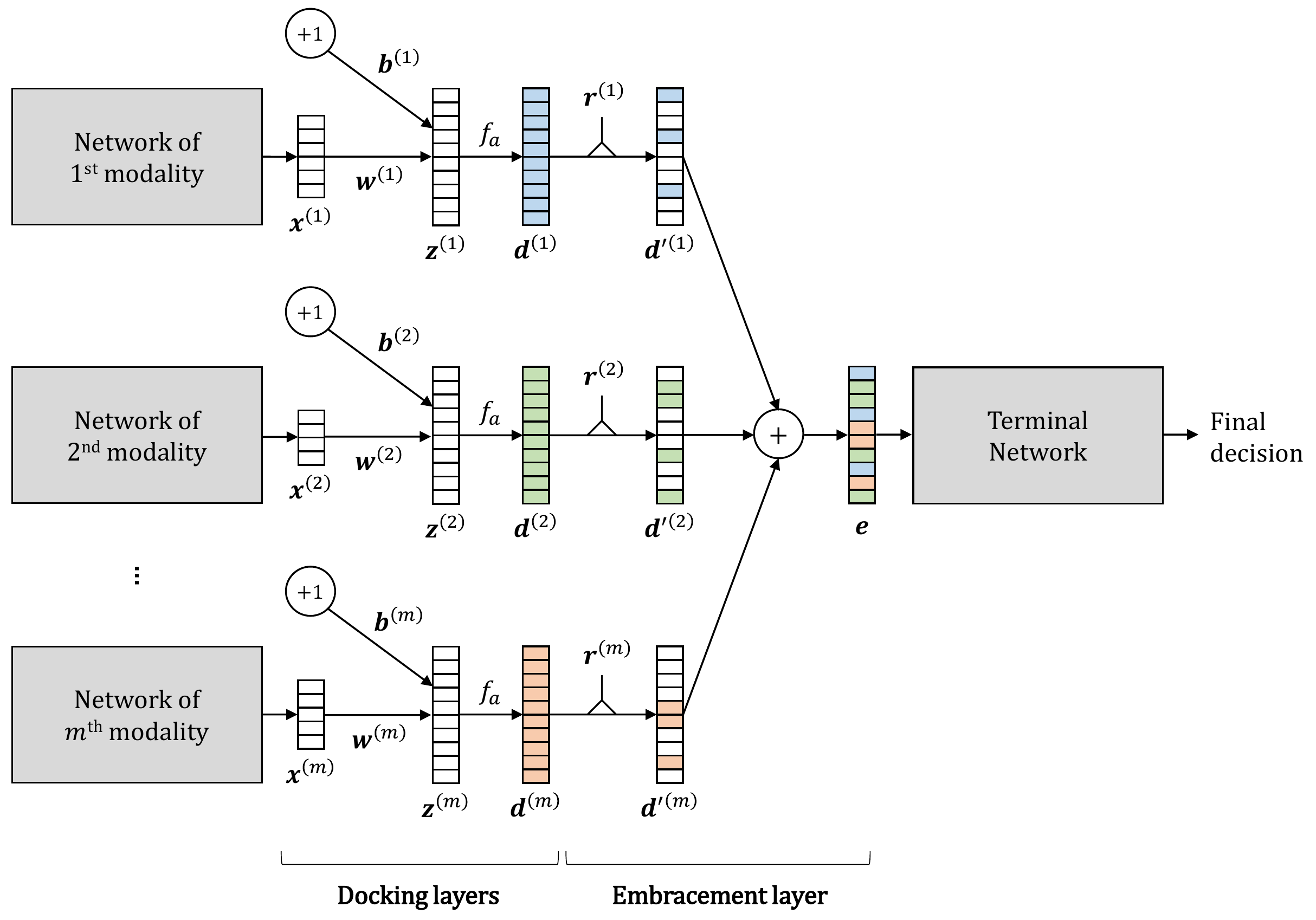}
	\caption{Overall structure of the proposed EmbraceNet model.}
	\label{fig:embracenet_structure}
\end{figure*}

Some researches employed data processing to handle missing data in deep learning architectures.
Ord{\'o}{\~n}ez and Roggen used a linear interpolation method to handle missing data for human activity recognition tasks \cite{ordonez2016deep}.
Ngiam \textit{et al.} trained their audio-visual bimodal deep network by feeding zero values in the case of missing data \cite{ngiam2011multimodal}.
Eitel \textit{et al.} analyzed the pattern of noise existing in the depth data and augmented the training data by the observed noise pattern for object recognition using color and depth images \cite{eitel2015multimodal}.
These methods may partially improve the robustness against missing data, however, only provide simple workarounds and do not solve the issue fundamentally via learning or modeling.
Jaques \textit{et al.} introduced an approach where a pre-trained autoencoder estimates the original values for missing part of the input data \cite{jaques2017multimodal}.
Nevertheless, the multimodal integration may fail if the estimation is not sufficiently accurate.

There are a few network models designed to cope with missing modalities.
Srivastava and Salakhutdinov built a deep Boltzmann machine (DBM) that learns shared representation from multimodal data and can disregard missing modalities in generating the joint representation via the Gibbs sampling method \cite{srivastava2012multimodal}.
Gu \textit{et al.} proposed a deep learning model containing multi-fusion layers, which have unimodal channels to learn modality-specific characteristics and a bimodal joint channel to learn correlated information \cite{gu2017learning}.
However, these are either limited to specific deep learning architectures (e.g., DBM) or specific numbers of modalities (e.g., bimodal).

\section{Model description}
\label{sec:modeldescription}

In this section, we describe the overall structure of the EmbraceNet model, which is depicted in
\figurename~\ref{fig:embracenet_structure}.
Our model mainly consists of two parts: docking layers and an embracement layer.

\subsection{Docking layers}

EmbraceNet takes output vectors of independent network models of different modalities as inputs.
Each network model may preprocess the data acquired from the corresponding sensor, and can be any kind of network structure such as conventional multilayer perceptrons, CNN-based deep learning architecture, hand-crafted feature vectors, or even raw data.
Since the modalities can have different characteristics, sizes of the vectors outputted from the network models can differ.
Thus, before aggregating them, EmbraceNet converts each vector to a \textit{dockable} vector so that the vectors have the same size.

Assume that there are $m$ modalities and the corresponding network models.
Let $\mathbf{x}^{(k)}$ be the output vector from the $k$-th network model, where $k \in \{1, 2, ..., m\}$.
Then, the $i$-th component of the input vector of the $k$-th docking layer is represented as

\begin{equation}
	\label{eq:docking_layer_input}
	{z}_{i}^{(k)} = \mathbf{w}_{i}^{(k)} \cdot \mathbf{x}^{(k)} + {b}_{i}^{(k)}
\end{equation}
where $\mathbf{w}_{i}^{(k)}$ and ${b}_{i}^{(k)}$ are a weight vector and a bias, respectively.
An activation function ${f}_{a}$ (e.g., rectified linear unit (ReLU), sigmoid, or hyperbolic tangent) is applied to ${z}_{i}^{(k)}$ to obtain the output of the $k$-th docking layer, i.e., 

\begin{equation}
	\label{eq:docking_layer_output}
	{d}_{i}^{(k)} = {f}_{a}({z}_{i}^{(k)})
\end{equation}
with $\mathbf{d}^{(k)} = [{d}_{1}^{(k)}, {d}_{2}^{(k)}, ..., {d}_{c}^{(k)}]^{T}$.
Note that all outputs of the docking layers (i.e., $\mathbf{d}^{(1)}, \mathbf{d}^{(2)}, ..., \mathbf{d}^{(m)}$) are $c$-dimensional vectors, i.e., $i \in \{1, 2, ..., c\}$.

\subsection{Embracement layer}
\label{sec:embracement_layer}

There are $m$ vectors obtained from the docking layers, where each vector consists of $c$ values.
In the embracement layer of our proposed model, these vectors are combined to a single vector consisting of $c$ values, which is so-called an ``embraced'' vector.
A plain solution to do this is the element-wise summation of the vectors, which is highly vulnerable to partially available data as aforementioned.
Instead, the EmbraceNet model employs an elaborate fusion technique based on a multinomial sampling as follows.

Let $\mathbf{r}_{i} = [{r}_{i}^{(1)}, {r}_{i}^{(2)}, ..., {r}_{i}^{(m)}]^{T}$ ($i=1, 2, ..., c$) be a vector that is drawn from a multinomial distribution, i.e.,

\begin{equation}
\label{eq:embrace_layer_toggle_selector}
\mathbf{r}_{i} \sim \textrm{Multinomial}(1, \mathbf{p})
\end{equation}
where $\mathbf{p} = [{p}_{1}, {p}_{2}, ..., {p}_{m}]^{T}$ and $\sum_{k}{p_k} = 1$.
This satisfies that only one value of $\mathbf{r}_{i}$ is equal to 1 and the rest values are equal to 0.
Then, the vector $\mathbf{r}^{(k)} = [{r}_{1}^{(k)}, {r}_{2}^{(k)}, ..., {r}_{c}^{(k)}]^{T}$ is applied to the vector $\mathbf{d}^{(k)}$ as

\begin{equation}
	\label{eq:toggled_docking_layout_output}
	\mathbf{d}^{\prime(k)} = [{d}_{1}^{\prime(k)}, {d}_{2}^{\prime(k)}, ..., {d}_{c}^{\prime(k)}]^{T} = \mathbf{r}^{(k)} \circ \mathbf{d}^{(k)}
\end{equation}
where $\circ$ denotes the Hadamard product (i.e., ${d}_{i}^{\prime(k)} = {r}_{i}^{(k)} \cdot {d}_{i}^{(k)}$).
Finally, the $i$-th component of the output vector of the embracement layer $\mathbf{e} = [{e}_{1}, {e}_{2}, ..., {e}_{c}]^{T}$ is obtained as

\begin{equation}
	\label{eq:embrace_layer_output}
	{e}_{i} = \sum_{k}{{d}_{i}^{\prime(k)}}.
\end{equation}
The length of $\mathbf{e}$ is the same to that of $\mathbf{d}^{(k)}$ (i.e., $c$).
It then serves as an input vector of the terminal network, which outputs a final decision of the given classification task.

The aforementioned process ensures that only one modality data contributes to each component of the embraced vector (i.e., ${e}_{i}$).
For example, if $\mathbf{r}_{4} = [1, 0, ..., 0]^{T}$, then ${e}_{4}={d}_{4}^{\prime(1)}$ as shown in \figurename~\ref{fig:embracenet_structure}.
Nevertheless, since this so-called ``modality selection'' process is performed independently in every component of $\textbf{e}$, the final output of the embracement layer is obtained from data of all modalities, thus the multimodal information is eventually integrated.
This dynamic process brings various advantages in real-world multimodal scenarios, which is discussed in Section~\ref{sec:benefits}.

The modality selection process is dependent on the probability values of $\mathbf{p}$.
Typically, $\mathbf{p} = [1/m, 1/m, ..., 1/m]^{T}$ can be used to give equal chances to all modalities during the modality selection.
In addition, we show how it can be further tuned for improved performance during the training phase in Section~\ref{sec:optimizing_training_stage}, and even after the training phase in Section~\ref{sec:optimizing_testing_stage}.

\section{Benefits of the EmbraceNet architecture}
\label{sec:benefits}

The docking and embracing structure of the EmbraceNet model provides a powerful solution to fuse multimodal information. 
We discuss three main benefits of the architecture in this section.

\subsection{Considering correlations between modalities}

If the multimodal data are acquired from the same target, there usually exist cross-modal correlations between the data obtained from different sensors.
For example, utilizing correlations between acoustic features and visual cues in speech recognition is beneficial to improve the overall performance \cite{lee2009two}.

The EmbraceNet properly considers correlations between modalities in the process of training the docking layers.
Once the docking layer of each modality produces its own representation $\textbf{d}^{(k)}$, only some parts of the vector are further processed in the embracement layer.
In addition, the indices of the reflected values randomly change at every training iteration.
Therefore, each docking layer tries to make its output similar to the representations calculated from the other docking layers, so that the embracement layer outputs the same representation for the same input.
This mechanism leads the docking layer for a modality to learn its weights not only from its own input data, but also from the characteristics of the other modalities.
We show that the EmbraceNet efficiently considers the correlations between different modalities in Section~\ref{sec:results}.

\subsection{Handling missing data}

In the wild, parts of data may be lost during acquisition or transmission of the data.
For example, wirelessly connected sensors can be occasionally disconnected so that the data cannot be obtained for a period of time \cite{chavarriaga2013opportunity}.
In addition, even if data loss does not happen, some technical errors may corrupt the data, which may need to be discarded \cite{higuera2015self,taamneh2017multimodal}.

EmbraceNet can robustly handle missing data at the embracement layer by adjusting the probabilities $\mathbf{p}$. 
Let $\mathbf{u} = [{u}_{1}, {u}_{2}, ..., {u}_{m}]^{T}$ be a vector describing the presence of each modality, where

\begin{equation}
\label{eq:presence_vector}
{u}_{k} = 
\begin{cases}
1 & \textrm{if } \mathbf{x}^{(k)} \textrm{ exists} \\
0 & \textrm{otherwise}
\end{cases}
.
\end{equation}
Then, we adjust the multinomial distribution of $s_i$ by altering $\mathbf{p}$ to $\mathbf{\hat{p}} = [\hat{p}_{1}, \hat{p}_{2}, ..., \hat{p}_{m}]^{T}$ where

\begin{equation}
\label{eq:p_vector_hat}
\hat{p}_{k} = \frac{{u}_{k} {p}_{k}}{\sum_{j}{{u}_{j}{p}_{j}}}.
\end{equation}
If the data of the $k$-th modality is not available, the value of ${u}_{k}$ becomes 0 and $\hat{p}_{k}$ becomes 0 accordingly.
This eliminates the chance that the value of ${r}_{i}^{(k)}$ becomes 1.
Therefore, the invalid data coming from the $k$-th network model does not propagate to the EmbraceNet model.
Instead, valid data of the other network models contribute more in generating the output representation of the embracement layer.

\subsection{Regularization effect}
\label{sec:regularization_effect}

The modality selection process explained in Section~\ref{sec:embracement_layer} chooses one of the docking layers to serve as an input of the embracement layer.
In the perspective of the $k$-th network model, each value of its docking layer is selected to be forwarded when ${r}_{i}^{(k)} = 1$, where the probability of ${r}_{i}^{(k)} = 1$ is given by ${p}_{k}$.
Therefore, the operation at the $k$-th docking layer satisfies a binomial distribution, i.e.,

\begin{equation}
\label{eq:each_model_binomial_distribution}
{r}_{i}^{(k)} \sim \textrm{Binomial}(1, {p}_{k})
.
\end{equation}
Since the binomial distribution with drawing one random variable is equivalent to the Bernoulli distribution, this process is equivalent to the dropout operation \cite{srivastava2014dropout}, where each value in the output vector is dropped out with probability $(1 - {p}_{k})$ for preventing overfitting via regularization.
Therefore, although the mechanism of the embracement layer is introduced for modality selection, it additionally results in effective prevention of excessive learning towards the data of specific modalities during the training phase.

\section{Optimizing the EmbraceNet architecture}
\label{sec:optimizing}

\begin{figure}[t]
	\centering
	\subfigure[]{
		\includegraphics[width=1.35in]{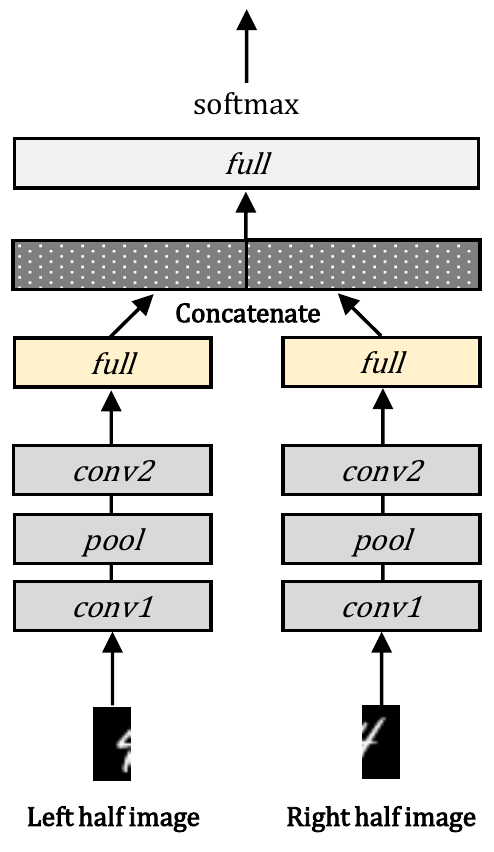}
	}
	\hspace{0.15in}
	\subfigure[]{
		\includegraphics[width=1.35in]{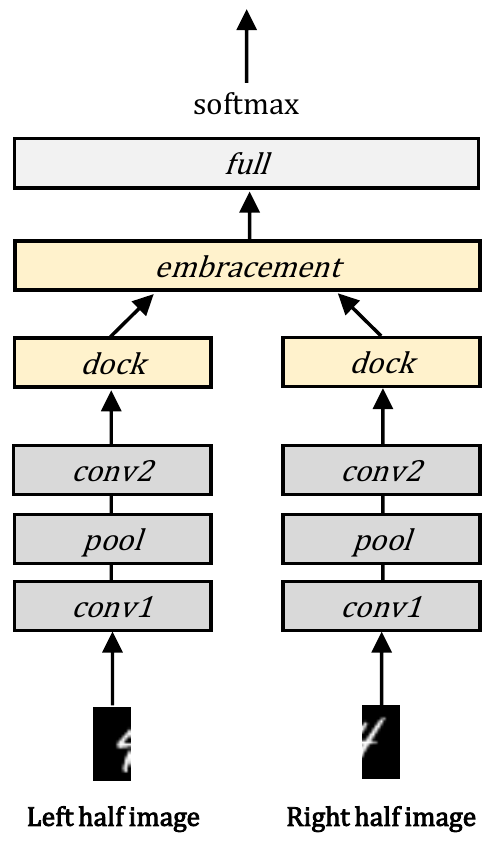}
	}
	\caption{Structures of the network models employed on the bimodal MNIST and Fashion MNIST datasets. (a) Intermediate integration-based network (b) EmbraceNet-based network}
	\label{fig:mnist_models}
\end{figure}

\begin{table*}
	\small
	\renewcommand{\arraystretch}{1.2}
	\centering
	\begin{tabular}{c|ccc|ccc}
		\hline
		\hline
		& \multicolumn{3}{c|}{Bimodal MNIST} & \multicolumn{3}{c}{Bimodal Fashion MNIST} \\
		\hline
		& ~All~ & Left only & Right only & ~All~ & Left only & Right only \\
		\hline
		Intermediate integration & 0.95 & 15.17 & 16.06 & 8.26 & 29.23 & 15.49 \\
		EmbraceNet (baseline) & 0.95 & 9.07 & 9.70 & 8.11 & 14.64 & 11.79 \\
		\hline
		EmbraceNet (+Adj. pre) & 0.95 & \textbf{3.35} & \textbf{4.31} & 8.03 & \textbf{9.80} & \textbf{9.66} \\
		EmbraceNet (+Adj. post) & \textbf{0.93} & 9.07 & 9.70 & 8.06 & 14.64 & 11.79 \\
		EmbraceNet (+Adj. both) & \textbf{0.93} & \textbf{3.35} & \textbf{4.31} & \textbf{7.97} & \textbf{9.80} & \textbf{9.66} \\
		\hline
		\hline
	\end{tabular}
	\caption{Error rates (\%) of the classification models on the bimodal MNIST and Fashion MNIST datasets.}
	\label{table:toy_experiment}
\end{table*}

The EmbraceNet architecture introduces new parameters denoted as $\textbf{p}$.
These parameters enable to robustly learn multimodal representations and regulate the activations of the network models in case of missing data.
Here, we show how the configurations of the parameters affect the overall performance of the EmbraceNet architecture.

To briefly show effectiveness of the optimization methods, we conduct toy experiments with the MNIST \cite{lecun1998gradient} and Fashion MNIST \cite{xiao2017fashion} datasets.
Each dataset contains 60000 training and 10000 test images having a size of $28 \times 28$ pixels.
We divide the images into left and right halves having a size of $14 \times 28$ pixels and considered as bimodal datasets.
A baseline intermediate integration-based and an EmbraceNet-based network having similar structures are constructed as shown in \figurename~\ref{fig:mnist_models}.
In both networks, the convolutional layers ($conv1$ and $conv2$) have 64 convolutional units each. A max pooling ($pool$) layer is used after $conv1$.
The intermediate integration-based network concatenates the output of the fully connected layers ($full$ after $conv2$) having 512 neurons each, and passes it to the final softmax layer ($full$).
On the other hand, the EmbraceNet-based network passes the outputs of the two $conv2$ layers to the EmbraceNet structure with $c=512$ and then to the final softmax layer ($full$).
All the networks are trained by the Adam optimization method \cite{kingma2014adam} with ${\beta}_{1}=0.9$, ${\beta}_{2}=0.999$, $\hat{\epsilon}={10}^{-2}$, a learning rate of $10^{-3}$, and a batch size of 64.
During the training of the baseline EmbraceNet-based network, both probabilities in $\textbf{p}$ are set to 0.5.

We train the models with 60000 bimodal samples and tested with all combinations of the modalities (i.e., left halves only, right halves only, and both halves) from 10000 samples.
For the intermediate integration, the simulated missing part of an image is replaced with zero-valued data.
The results are shown in Table~\ref{table:toy_experiment}.
The top two rows of Table~\ref{table:toy_experiment} show the error rates of the baseline models.
They show the similar performance when no missing data is introduced (the columns for ``All'').
However, when only one of the two halves is given (the columns for ``Left only'' and ``Right only''), the EmbraceNet significantly outperforms the intermediate integration model, despite the lack of significant information for recognition.

\subsection{Adjusting parameters during training}
\label{sec:optimizing_training_stage}

The original dropout mechanism \cite{srivastava2014dropout} is basically motivated by the theory of sexual reproduction \cite{livnat2010sex}.
According to the theory, a set of genes having higher ability to cooperate with another random set is more robust against a long run of the natural selection process.
If the set of genes is highly dependent on only specific sets of genes, it may be hard to survive when new sets are introduced.
In the perspective of artificial neural networks, training with the dropout mechanism can make each neuron more robust to new samples and prevent the neuron from relying on only specific neurons.

As discussed in Section~\ref{sec:regularization_effect}, using $\mathbf{p}$ in the proposed architecture has a regularization effect equivalent to the dropout mechanism.
On top of this, the regularization effect can be more emphasized by controlling the probabilities in $\textbf{p}$.
For example, we can choose some of ${p}_{k}$ randomly at every training iteration and make them 0.
This process can block the propagation of specific modalities entirely, which avoids the embracement layer to rely on specific modalities.
While the existence of $\textbf{p}$ provides the dropout mechanism to each neuron of the embracement layer, adjusting the values of ${p}_{k}$ as above realizes the dropout mechanism to the embracement layer itself.

To show the effectiveness of this idea, we train the EmbraceNet models with only one modality randomly selected with a probability of 0.5 at every training iteration.
The classification results of the models are shown in the row denoted as ``EmbraceNet (+Adj. pre)'' in Table~\ref{table:toy_experiment}.
While the error rate for non-missing data is kept unchanged, the error rates for missing data are significantly reduced compared to those of the baseline EmbraceNet models.
This confirms that the advanced regularization can significantly improve the performance with respect to different combinations of the modalities.

\subsection{Adjusting parameters after training}
\label{sec:optimizing_testing_stage}

Since various types of modalities can be involved in the information fusion, the importance of each modality can differ.
For example, in the top two rows of Table~\ref{table:toy_experiment}, the error rates in Bimodal MNIST for the right-side images are higher than those for the left-side images.
On the other hand, the error rates in Bimodal Fashion MNIST for the left-side images are higher than those for the right-side images.
These imply that the amount of information is not the same in the left and right parts of the images.

In the EmbraceNet architecture, controlling the contribution of each modality to the classification result can be easily done without any modification of the structure.
While the probabilities in $\textbf{p}$ play a role to regularize activations during training, they can be also used as parameters to maximize the performance of an already trained model.
During the testing stage, the probability values of $\textbf{p}$ control the proportions of the activation values of the docking layers to be embraced in the embracement layer.
Hence, changing the values can allow the model to incorporate relative reliabilities of different modalities.

To prove this, we assume that the performance of each modality on the training data provides us its relative reliability and use this information for adjusting the parameters ${p}_{k}$.
Therefore, the accuracies of each trained baseline EmbraceNet model are measured for left-only and right-only images of the 60000 training images, and the values of ${p}_{k}$ during the testing stage are set as the proportions of these accuracies.
On the bimodal MNIST dataset, the training accuracies are measured as 90.89\% and 90.92\%, respectively.
On the bimodal Fashion MNIST dataset, they are measured as 88.91\% and 92.51\%, respectively.
Note that the test images are not used to determine the values of ${p}_{k}$.
The row denoted as ``EmbraceNet (+Adj. post)'' in Table~\ref{table:toy_experiment} shows the performance obtained in this way.
It can be observed that in comparison to the baselines, the error rate is lowered from 0.95\% to 0.93\% on the bimodal MNIST dataset and from 8.11\% to 8.06\% on the bimodal Fashion MNIST dataset.
Note that the error rates for missing data remain the same since $\textbf{p}$ does not have any effect when only a single modality is available.

Furthermore, the model trained with the optimized probabilities in $\textbf{p}$ (i.e., ``EmbraceNet (+Adj. pre)'') is also evaluated with the adjusted $\textbf{p}$.
The new values of ${p}_{k}$ for testing are set as the proportions of the accuracy for each modality measured on the training images.
They are 98.20\% and 97.53\%, respectively, for the bimodal MNIST dataset, and 94.98\% and 94.91\%, respectively, for the bimodal Fashion MNIST dataset.
The results are shown in the last row of Table~\ref{table:toy_experiment} (i.e., ``EmbraceNet (+Adj. both)'').
It shows that the model optimized both during and after the training stage performs better than the other models for all combinations of the modalities.

\begin{figure*}[t]
	\centering
	\subfigure[]{
		\includegraphics[width=0.145\linewidth]{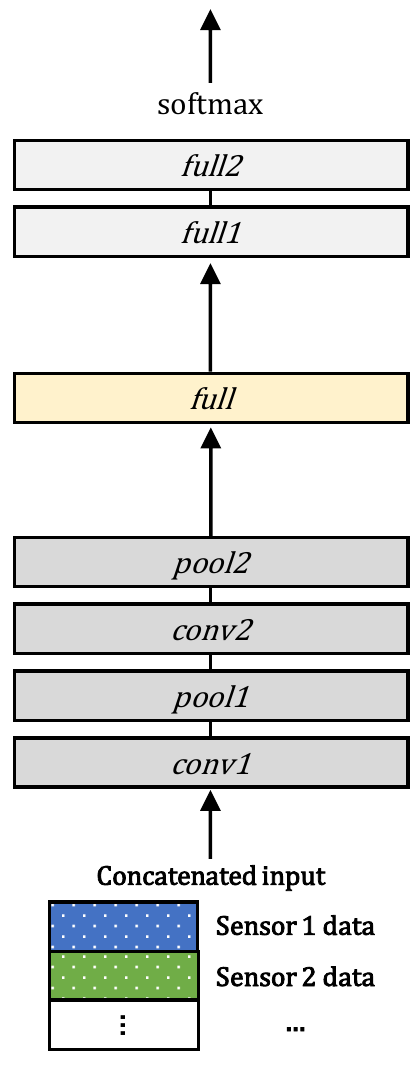}
	}
	\hspace{0.001\linewidth}
	\subfigure[]{
		\includegraphics[width=0.145\linewidth]{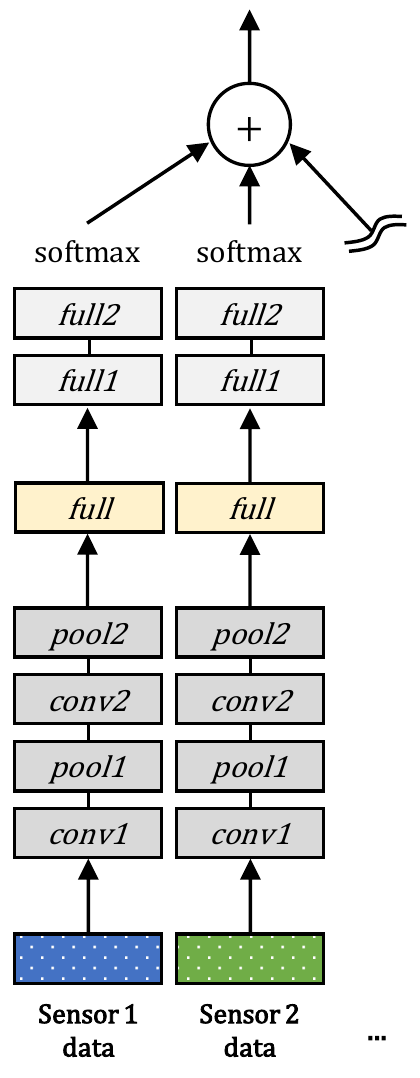}
	}
	\hspace{0.001\linewidth}
	\subfigure[]{
		\includegraphics[width=0.145\linewidth]{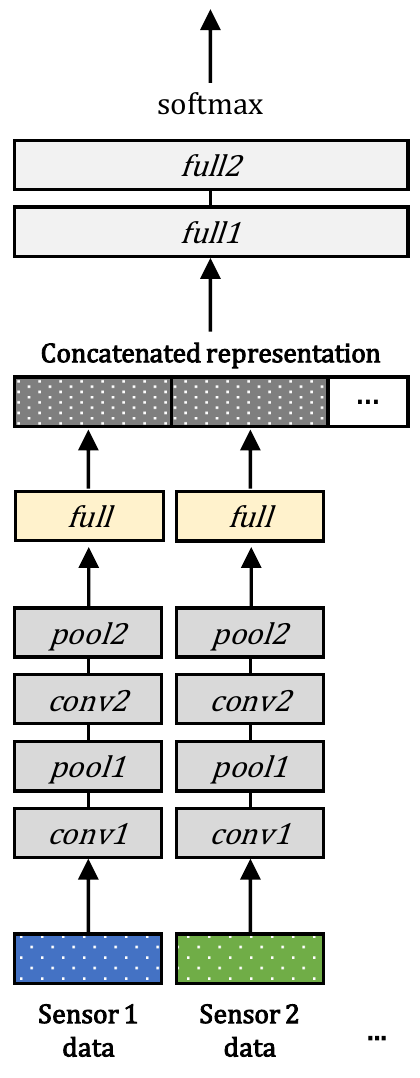}
	}
	\hspace{0.001\linewidth}
	\subfigure[]{
		\includegraphics[width=0.145\linewidth]{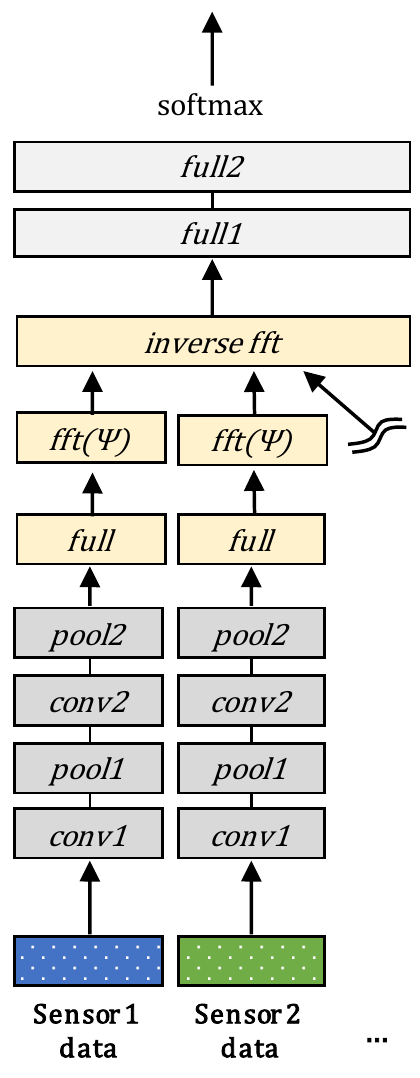}
	}
	\hspace{0.001\linewidth}
	\subfigure[]{
		\includegraphics[width=0.145\linewidth]{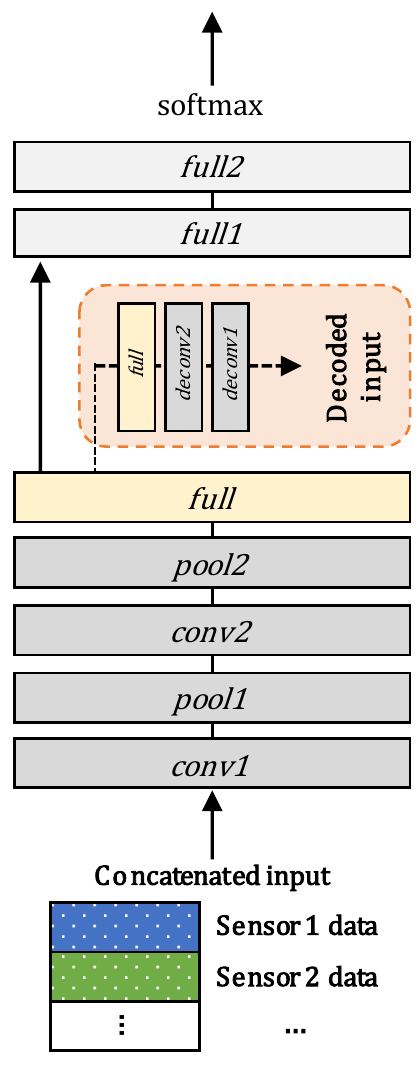}
	}
	\hspace{0.001\linewidth}
	\subfigure[]{
		\includegraphics[width=0.145\linewidth]{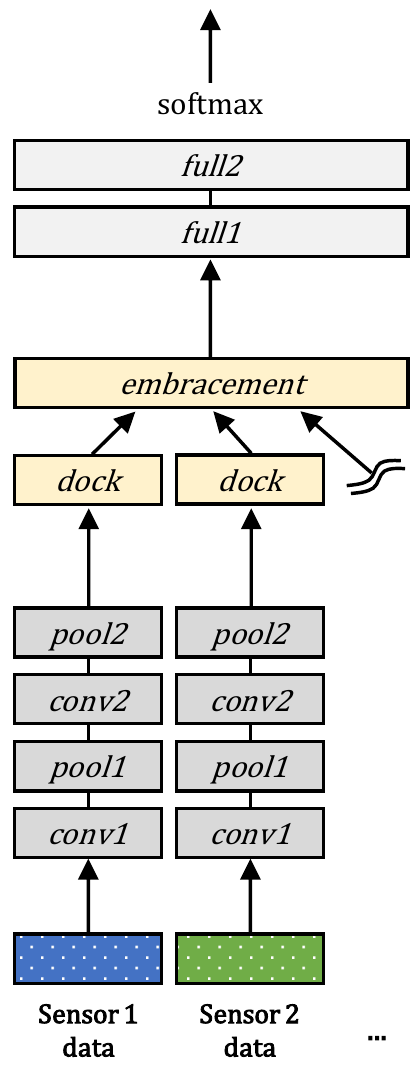}
	}
	\caption{Structures of the employed network models for the gas sensor arrays dataset. (a) Early integration (b) Late integration (c) Intermediate integration (d) Compact multi-linear pooling \cite{algashaam2017multispectral} (e) Multimodal autoencoder \cite{jaques2017multimodal} (f) EmbraceNet}
	\label{fig:experiment_models_gas}
\end{figure*}

\begin{figure*}[t]
	\centering
	\subfigure[]{
		\includegraphics[width=0.145\linewidth]{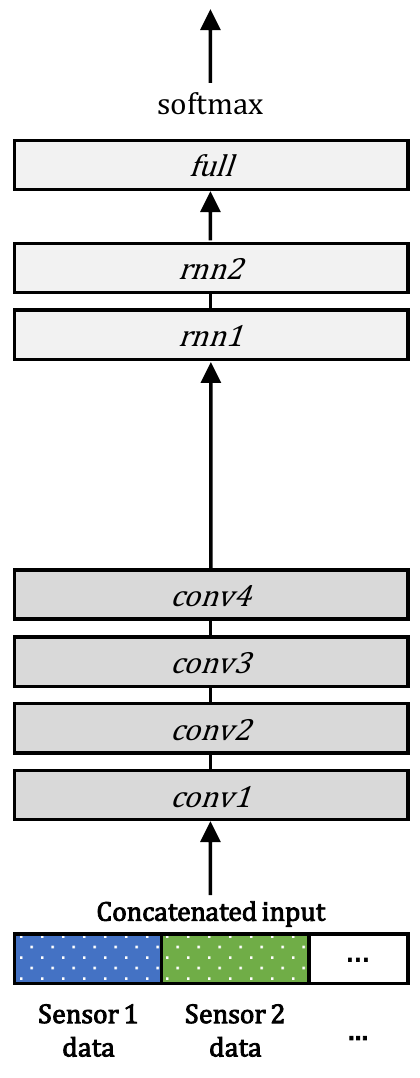}
	}
	\hspace{0.001\linewidth}
	\subfigure[]{
		\includegraphics[width=0.145\linewidth]{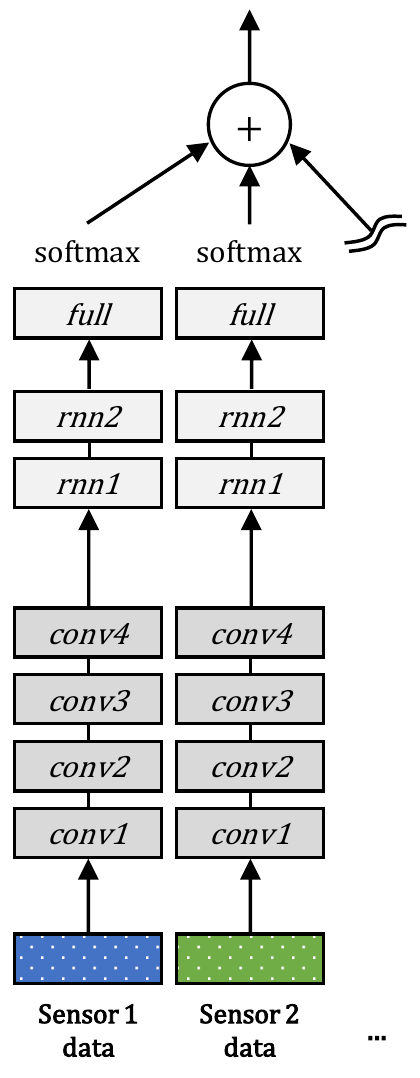}
	}
	\hspace{0.001\linewidth}
	\subfigure[]{
		\includegraphics[width=0.145\linewidth]{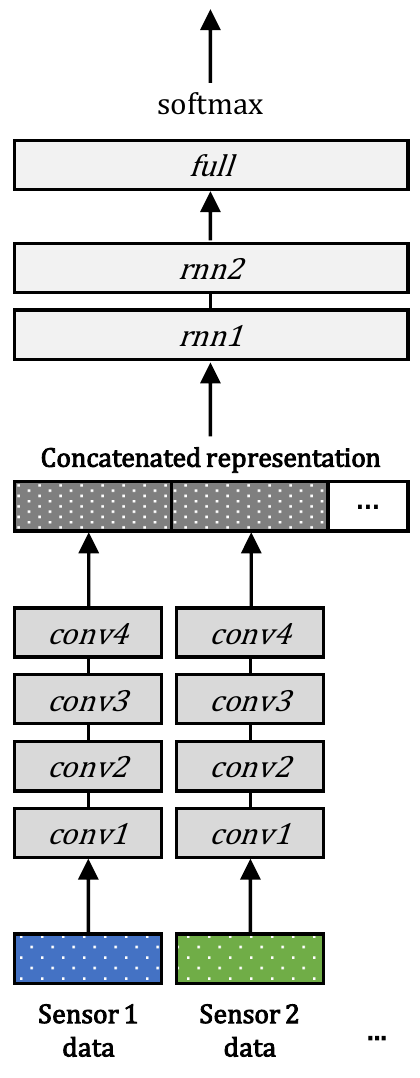}
	}
	\hspace{0.001\linewidth}
	\subfigure[]{
		\includegraphics[width=0.145\linewidth]{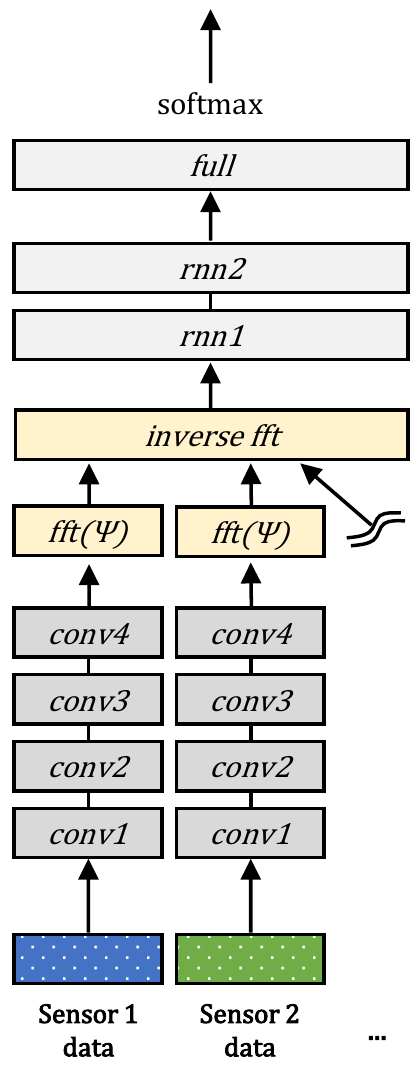}
	}
	\hspace{0.001\linewidth}
	\subfigure[]{
		\includegraphics[width=0.145\linewidth]{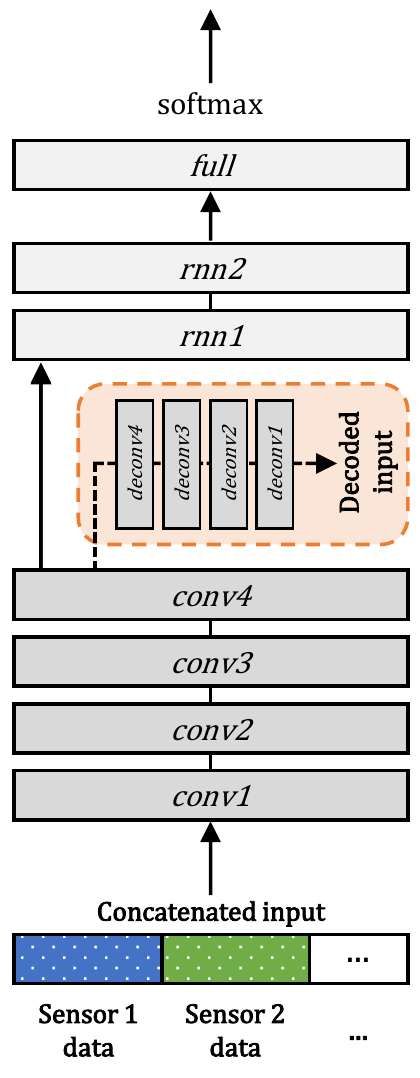}
	}
	\hspace{0.001\linewidth}
	\subfigure[]{
		\includegraphics[width=0.145\linewidth]{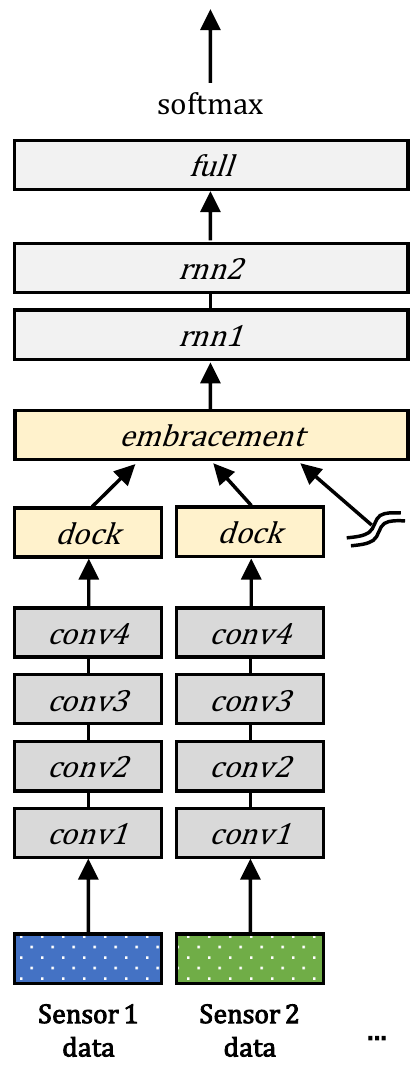}
	}
	\caption{Structures of the employed network models for the OPPORTUNITY dataset. (a) Early integration (b) Late integration (c) Intermediate integration (d) Compact multi-linear pooling \cite{algashaam2017multispectral} (e) Multimodal autoencoder \cite{jaques2017multimodal} (f) EmbraceNet}
	\label{fig:experiment_models_opportunity}
\end{figure*}

\section{Comparison with other multimodal fusion techniques}
\label{sec:setup}

We conduct experiments to evaluate the effectiveness of the EmbraceNet model by comparing with other classification models.
This section provides details of the datasets, the network models, and the experimental setups.

\subsection{Datasets}

We employ two multimodal datasets for classification: the gas sensor arrays dataset \cite{vergara2013performance} for classifying chemical sources and the OPPORTUNITY dataset \cite{roggen2010collecting} for recognizing human activities.

\subsubsection{Gas sensor arrays dataset}

The gas sensor arrays dataset \cite{vergara2013performance} contains 18000 sensor measurement sequences obtained from 10 chemical sources (e.g., \textit{acetone}, \textit{butanol}) and 72 gas sensors.
The gas sensors were placed on nine sensor arrays, where each sensor array consists of eight different metal-oxide gas sensors.
A specially designed wind tunnel test bed was built to obtain the dataset.
A chemical source was placed at one end of the tunnel and a fan attached at the other end of the tunnel blew the gas towards outside the tunnel.
The sensor arrays were placed between the chemical source and fan and measured electrical conduction through the metal-oxide film of the sensors for 260 seconds with a sampling rate of 100 Hz.
The dataset acquisition was performed with various configurations, including three wind speeds of the fan, five different temperatures of sensors, and six different locations of the sensor arrays.

In our experiments, we consider eight modalities that correspond to different types of gas sensors.
We downsample the data to 1 Hz and extract multiple sets of recordings from each sequence in the training dataset by windowing with a step size of one second to augment the number of data.
Hence, a sensor data with a size of $240 \times 9$ is given for each modality.
All the sensor data are normalized to have values within $[0.0, 1.0]$.
We employ the data obtained from the first, third, and fifth locations of the sensor arrays for training, the second location for validation, and the fourth location for testing\footnote{The sixth location is excluded because the data for \textit{butanol} on that location are entirely missing in the original dataset.}.

\subsubsection{OPPORTUNITY dataset}

The OPPORTUNITY activity recognition dataset \cite{roggen2010collecting} consists of multimodal data recorded from four subjects.
During the data acquisition, seven inertial sensors and 12 accelerometers were attached on the subjects.
The dataset consists of five free activity sequences (i.e., activity of daily living (ADL) sessions) and a controlled sequence (i.e., drill session) for each subject.
In an ADL session, the subject freely performed under given activity instructions (e.g., preparing coffee, eating sandwich, cleanup).
In a drill session, the subject performed 17 activities in a predefined order with 20 repetitions.
All sensor data were synchronized and acquired with a sampling rate of 30 Hz.

There was an activity recognition challenge based on the OPPORTUNITY dataset \cite{chavarriaga2013opportunity}.
Among the four tasks in the challenge, we conduct the multimodal recognition of the modes of locomotion and mid-level gestures.
The modes of locomotion represent four classes, including \textit{stand}, \textit{walk}, \textit{sit}, and \textit{lie}.
The gestures represent 17 activities, including \textit{open/close door}, \textit{open/close fridge}, \textit{open/close dishwasher}, \textit{open/close drawer}, \textit{clean table}, \textit{drink from cup}, and \textit{toggle switch}.
Each task also contains the \textit{null} class, which indicates that there is no obvious activity in that time.
By following the protocol of the OPPORTUNITY challenge, a total of 19 body-worn sensors having 113 sensor channels are used (i.e., $m=19$).

We also follow the protocol of the challenge for splitting the dataset.
For training the network models, all sessions of subject 1, the first two ADL sessions of subjects 2 and 3, and the drill sessions of subjects 2 and 3 are used.
For testing, the last two ADL sessions of subjects 2 and 3 are used.
The third ADL sessions of subjects 2 and 3 are used for validation, where the data for the \textit{null} class in classifying the gestures are excluded to reduce the class imbalance of the data.
For data preprocessing, all sensor channels are normalized to have values within $[-1.0, 1.0]$.
Then, the data sequences are segmented as specified in the protocol of the OPPORTUNITY challenge.
A sliding window having a length of 500 ms (corresponding to 15 samples) is used, whose step size is set to 250 ms for the validation and test data and 33.3 ms for the training data.

\subsection{Performance measure}

To measure classification performance, we employ the weighted $F_1$ score as in \cite{ordonez2016deep} and \cite{chavarriaga2013opportunity}.
It is calculated as

\begin{equation}
\label{eq:weighted_f1_score}
{F}_{1} = \sum_{i}{2 \times \frac{{n}_{i}}{n} \times \frac{{precision}_{i} \times {recall}_{i}}{{precision}_{i} + {recall}_{i}}}
\end{equation}
where $n_i$ is the number of data in the $i$-th class, $n$ is the total number of data, ${precision}_i$ is the precision of the $i$-th class, and ${recall}_i$ is the recall of the $i$-th class.
The term ${{n}_{i}}/{n}$ balances the performance score against a skewed distribution of the classes.

\subsection{Network models}

We build five baseline deep learning network models for multimodal classification tasks.
The first three are basic multimodal integration models, which are early integration, late integration, and intermediate integration.
The rest two are an intermediate integration model with compact multi-linear pooling \cite{algashaam2017multispectral} and an early integration model pretrained by a multimodal autoencoder \cite{jaques2017multimodal}.
To ensure fair comparison, we design the network models to have the same structure except for the part for integration.
The overall structures of the network models employed for the gas sensor arrays and OPPORTUNITY datasets are illustrated in Figures~\ref{fig:experiment_models_gas} and \ref{fig:experiment_models_opportunity}, respectively.

For the gas sensor arrays dataset, the models consist of two convolutional layers ($conv1$ and $conv2$) with two max pooling layers ($pool1$ and $pool2$) and two fully connected layers ($full1$ and $full2$).
The first and second convolutional layers consist of 64 and 128 2-D convolutional units, respectively.
Each convolutional unit has a kernel size of $5\times5$ pixels.
The ReLU function is used as the activation function for the convolutional layers.
In addition, the max pooling layer reduces the size of activation values by factors of 2 and 3 in time and number of sensors, respectively.
For example, the size of output values of $conv1$ (i.e., $240\times9\times64$) is reduced to $120\times3\times64$.
The first fully connected layer ($full1$) outputs a vector having a length of $256$ and the second fully connected layer ($full2$) reduces its length to $10$ (i.e., the number of classes).

For the OPPORTUNITY dataset, we borrow an early integration-based model, ``DeepConvLSTM'', which was used for classifying activities in the OPPORTUNITY dataset and showed better performance than other methods \cite{ordonez2016deep}.
It consists of four convolutional layers ($conv1$, $conv2$, $conv3$, and $conv4$), two recurrent layers ($rnn1$ and $rnn2$), and one softmax dense layer ($full$).
The ReLU function is used as the activation function for the convolutional layers.
As the name of the model implies, long short-term memory (LSTM) cells \cite{hochreiter1997long} are used in the recurrent layers.
Each convolutional layer consists of 64 1-D convolutional units and each recurrent layer consists of 128 LSTM units.
The final softmax layer converts the output of the last recurrent layer ($rnn2$) for the last time step to probabilities for the target activities as in \cite{ordonez2016deep}.
The only difference between the original and our adapted models is the kernel size of the convolutional units: We use 3 instead of 5, because it is not possible to obtain valid representations from the data when the kernel size is 5\footnote{A sequence is required to have a length of at least 17 in order to be properly processed through four convolutional layers having a kernel size of 5, while each given data sequence has a length of 15 in the protocol of the OPPORTUNITY challenge.}.

\subsubsection{Early integration}

\figurename~\ref{fig:experiment_models_gas}~(a) shows the structure of the early integration model for the gas sensor arrays dataset.
We concatenate the data of the eight modalities to obtain the data having a size of $240\times72$.
This input data goes through two convolutional and max pooling layers, and the final max pooling layer produces its output having a size of $60\times3$.
Before we feed it into the two fully connected layers, another fully connected layer that outputs a vector having a length of $1024$ is employed.
The final output is converted to a probability distribution via a softmax function.

\figurename~\ref{fig:experiment_models_opportunity}~(a) shows the structure of the early integration model for the OPPORTUNITY dataset.
In this model, the shape of the input is $113\times15$, since there are 113 sensor channels.
The last convolutional layer ($conv4$) produces a representation with a size of $113 \times 64 \times 7$.
It is then reshaped to have a size of $7232 \times 7$, split to seven separate representations having a length of 7232, and fed into the recurrent layers that produce seven representations with a size of $128$.

\subsubsection{Late integration}

To ensure the consistency with the other network models, we duplicate the same structure of the early integration model for each modality, as shown in Figures~\ref{fig:experiment_models_gas} (b) and \ref{fig:experiment_models_opportunity} (b).
Therefore, there are eight and 19 separate networks for the gas sensor arrays and OPPORTUNITY datasets, respectively, where each network is dedicated to each modality.
After the training process, the $F_1$ score of each modality is measured for the validation dataset.
The calculated $F_1$ scores are used as the weights for integrating the class probabilities obtained at the final softmax layers of the unimodal network models via the weighted sum rule.
The final decision is determined by choosing the class having the maximum value among the aggregated class probabilities.

\subsubsection{Intermediate integration}

Figures~\ref{fig:experiment_models_gas}~(c) and \ref{fig:experiment_models_opportunity}~(c) show the intermediate integration models for the two datasets, respectively.
Since the convolutional units can be regarded as feature extractors \cite{choi2017impact,donahue2015long}, we employ the representation fusion step (i.e., concatenation) after the final convolutional layers.
As shown in the figures, the part before concatenation has the same structure to the lower part of the late integration model (until $pool2$ in \figurename~\ref{fig:experiment_models_gas} (b) and $conv4$ in \figurename~\ref{fig:experiment_models_opportunity} (b)), and the part after concatenation is the same to the upper part of the early integration model (from $full1$ in \figurename~\ref{fig:experiment_models_gas} (a) and $rnn1$ in \figurename~\ref{fig:experiment_models_opportunity} (a)).
While all the network structures and the number of units remain the same as the other network models, the output size of $full$ in \figurename~\ref{fig:experiment_models_gas} (c) is set to $128$ to prevent an excessively high dimension of the concatenated representation.
Therefore, the length of the concatenated representation in the middle is $128\times8 = 1024$ and $113\times64\times7 = 50624$ for the two datasets, respectively.


\begin{figure}[t]
	\centering
	\includegraphics[width=3.2in]{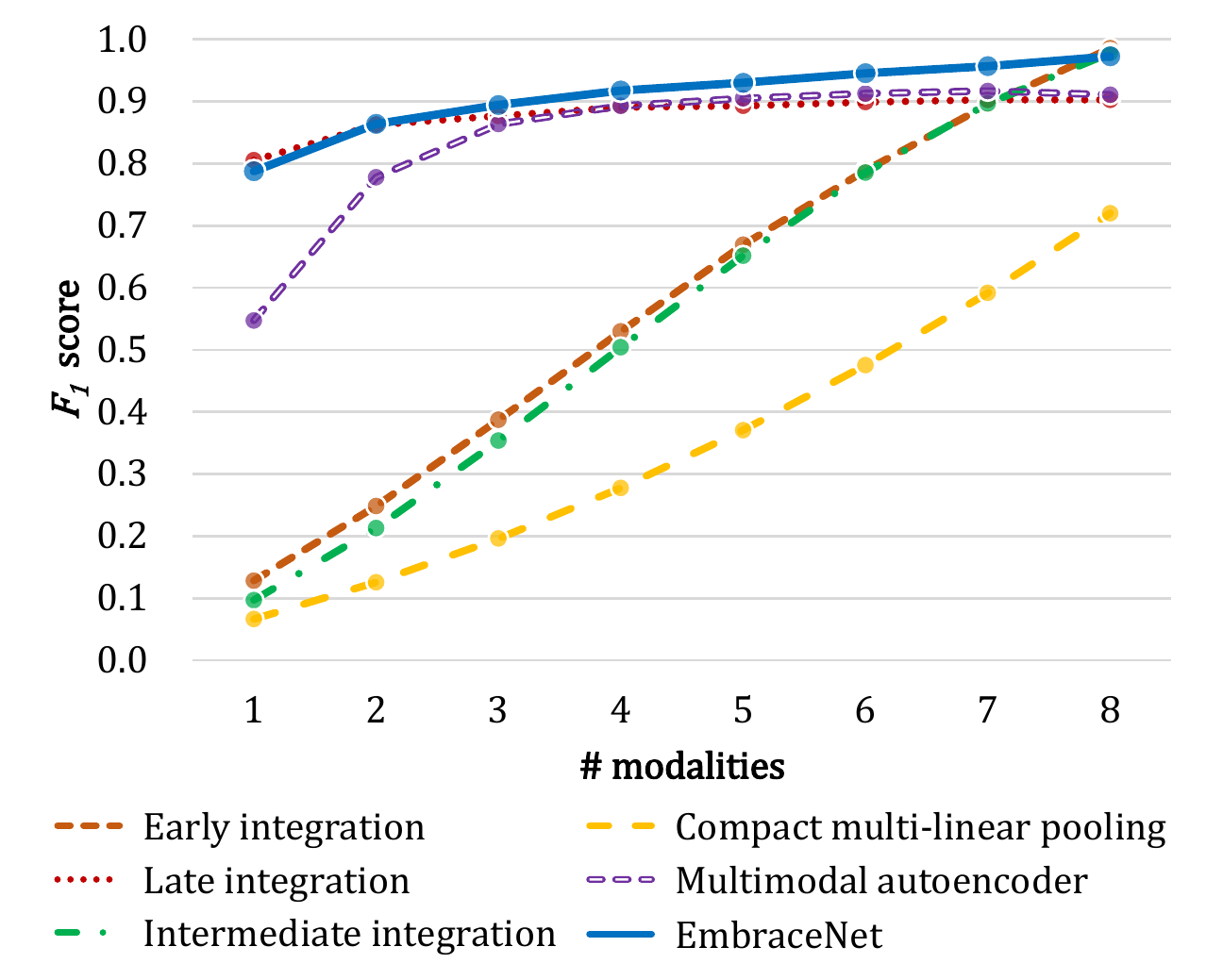}
	\caption{$F_1$ scores of the six models with respect to different numbers of modalities for classifying the gases in the gas sensor arrays dataset.}
	\label{fig:result_sensormissing_gas}
\end{figure}

\begin{figure*}[t]
	\centering
	\subfigure[]{
		\includegraphics[width=3.2in]{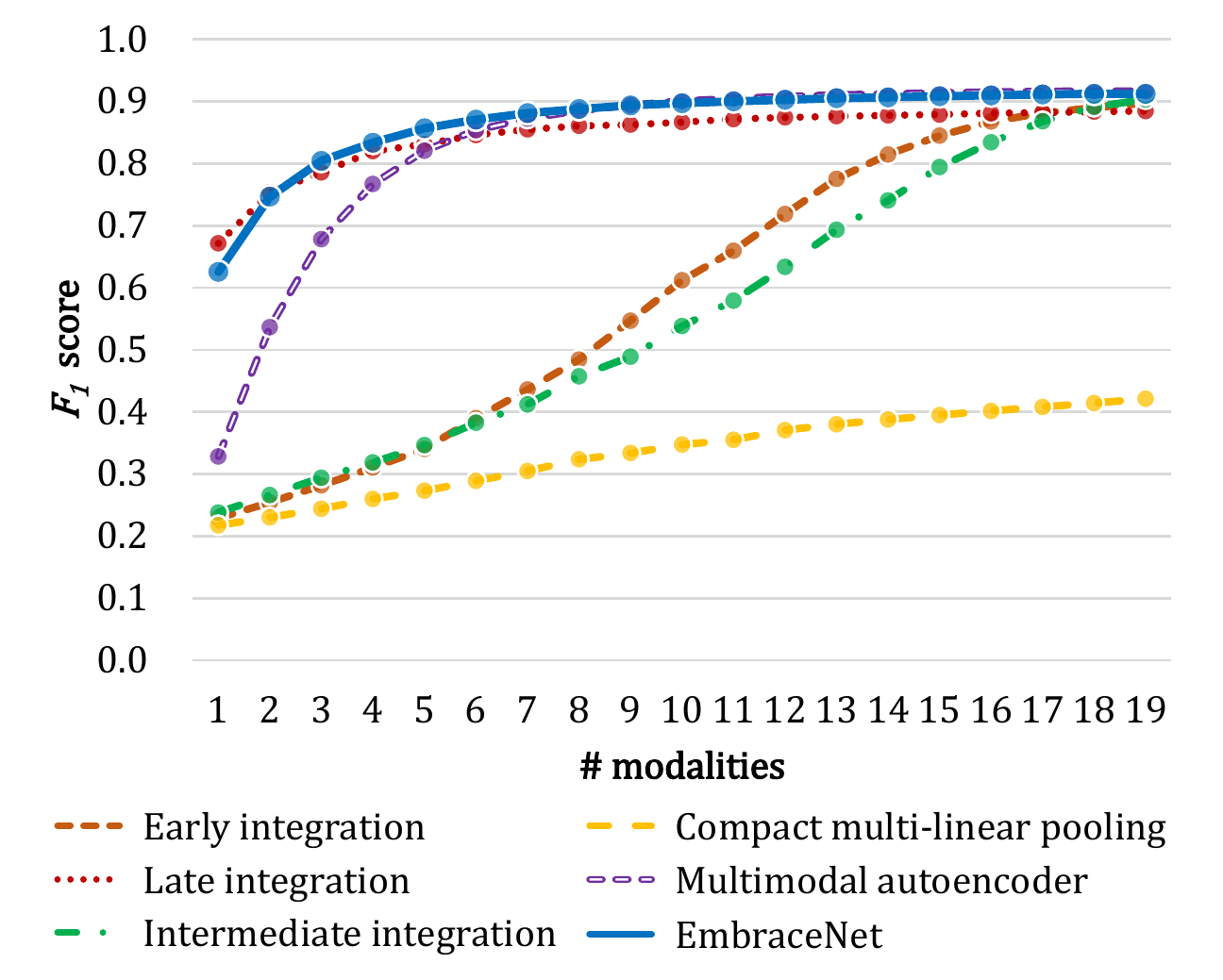}
	}
	\hspace{0.1in}
	\subfigure[]{
		\includegraphics[width=3.2in]{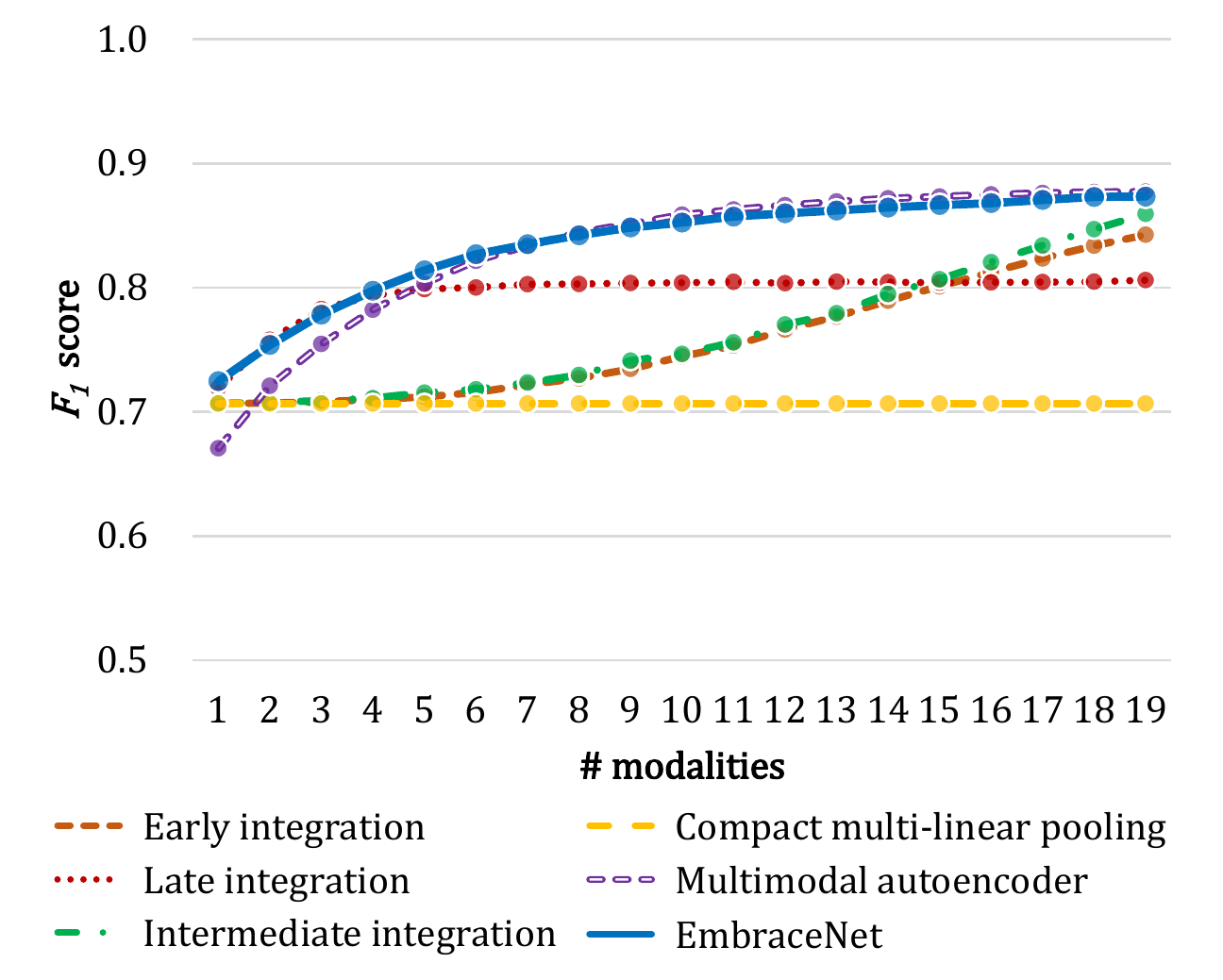}
	}
	\caption{$F_1$ scores of the six models with respect to different numbers of modalities for the classification tasks of the OPPORTUNITY dataset. (a) Modes of locomotion (b) Gestures}
	\label{fig:result_sensormissing_opportunity}
\end{figure*}

\subsubsection{Compact multi-linear pooling}

We employ the compact multi-linear pooling method \cite{gao2016compact,algashaam2017multispectral}.
As shown in Figures \ref{fig:experiment_models_gas} (d) and \ref{fig:experiment_models_opportunity} (d), the structures of the models are essentially the same to those of the intermediate integration models, except for the integration part.
By following the previous work \cite{gao2016compact,algashaam2017multispectral}, each model first calculates the count sketch \cite{pham2013fast} for each modality feature ($\Psi$ in Figures \ref{fig:experiment_models_gas} (d) and \ref{fig:experiment_models_opportunity} (d)), an element-wise multiplication is performed to the Fourier-transformed outputs of the count sketches, and the inverse Fourier transform is applied to the multiplied result.
In addition, a L2 normalization is applied to each modality before calculating its count sketch, as in the previous work \cite{gao2016compact,fukui2016multimodal}.

\subsubsection{Multimodal autoencoder}

We consider a multimodal autoencoder-based integration approach \cite{jaques2017multimodal} as another baseline, as shown in Figures \ref{fig:experiment_models_gas} (e) and \ref{fig:experiment_models_opportunity} (e).
The method concatenates the input data at the early stage.
Before training a classifier, the method first builds an autoencoder that is trained to reconstruct the original input data.
We set the encoding part of the autoencoder to have the same structure as the other baseline models.
For the decoding part, we employ the same number of fully connected layers and convolutional layers as in the encoding part, but in the reversed order and transposed manner.
We follow the procedure in \cite{jaques2017multimodal} for preprocessing the input data: We normalize the data to have values within $[0.0, 1.0]$ and replace the data of randomly selected 50\% of the modalities with a constant value of -1, which is not in the range of valid data values.
The cross-entropy function is used to calculate the reconstruction loss as in \cite{jaques2017multimodal}.
After the autoencoder is trained, we replace the decoding part with the classification layers, whose structure is the same to those of the other baselines.
Then, we train the model in an end-to-end manner, where the aforementioned procedure of preprocessing the input data is applied.

\subsubsection{EmbraceNet}

As shown in Figures \ref{fig:experiment_models_gas} (f) and \ref{fig:experiment_models_opportunity} (f), the overall structures are similar to those of the intermediate integration models.
The difference exists in treating representations obtained from the convolutional and max pooling layers.
We set $c = 1024$, thus the docking layers produce eight (for the gas sensor arrays dataset) or 19 (for the OPPORTUNITY dataset) output vectors having a length of $1024$, and the embracement layer combines them into one.
In the embracement layer, we give the same chance to be embraced for all the sensor data by setting all ${p}_{k}$ values to $1/m$ during the training process, where $m = 8$ and $m = 19$ for the gas sensor arrays and OPPORTUNITY datasets, respectively.
We apply the training procedure explained in Section~\ref{sec:optimizing_training_stage} by simulating data loss with an average modality missing probability of 50\%.
After the training, the network model is evaluated for the validation set, and the values of ${p}_{k}$ are determined according to the ratio of the $F_1$ scores, as explained in Section~\ref{sec:optimizing_testing_stage}.

\subsection{Scenarios}

\subsubsection{Missing modalities}

We first consider the scenario where the entire data of some sensors are missing.
This simulates the sensors that exist during training but are no longer available after deploying the trained model.
For the given test dataset, the performance is evaluated by varying combinations of the modalities.
For the gas sensor arrays dataset, all modality combinations are evaluated, where the total number of the combinations is $\sum_{i=1}^{8}{{8}\choose{i}} = 255$.
For the OPPORTUNITY dataset, because of too many possible combinations, at most 1000 randomly chosen modality combinations for each number of chosen modalities are evaluated.
Therefore, the total number of evaluated modality combinations is 14319, which is calculated by

\begin{equation}
	\sum_{i=1}^{19}{\min \bigg( 1000, {{19}\choose{i}} \bigg) }.
\end{equation}

Since the early integration, intermediate integration, and compact multi-linear pooling models require valid input data for all sensors, the values of a missing modality are replaced with the average value of the whole training data of the corresponding modality.
In addition, the test data that have missing parts in the original dataset are not used in order to accurately control the number of employed modalities.

\begin{figure*}[t]
	\centering
	\includegraphics[width=6.0in]{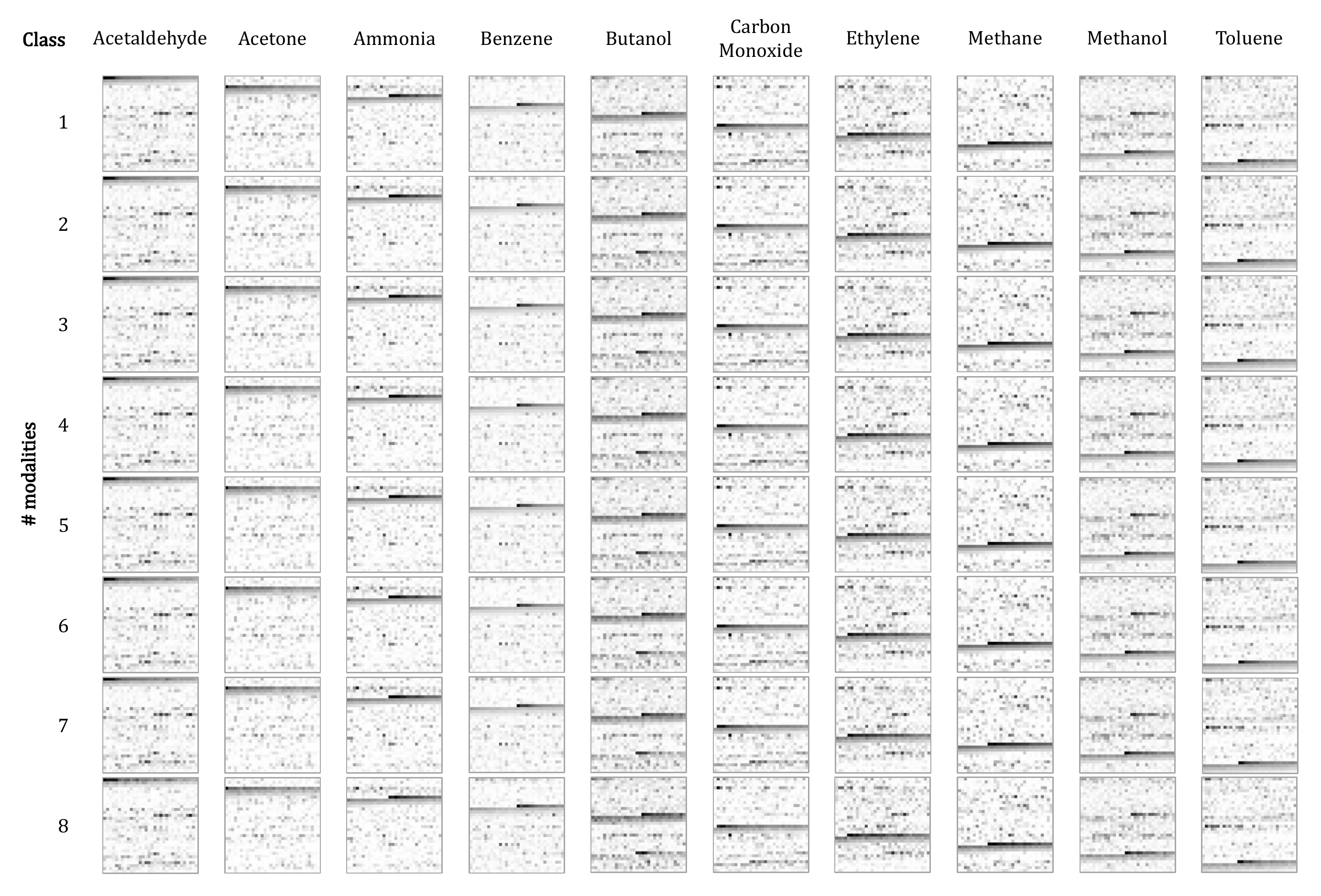}
	\caption{Averaged activations of the embracement layer with respect to different classes and numbers of modalities in the gas sensor arrays dataset. All the activations are sorted in the same order. A darker color means a higher value.}
	\label{fig:result_embraced_representations_gas}
\end{figure*}

\subsubsection{Missing block-wise data}

For the OPPORTUNITY dataset, we also consider the scenario where part of a data stream is lost.
To simulate this, we randomly choose one of the modalities and remove a randomly selected range of the data having a length between 300 and 900 time steps (i.e., 10 and 30 seconds).
The removing process is repeated until the desired missing rate is obtained, where the missing rate is calculated by measuring the number of missed time steps divided by the total number of data.
We use 10\% to 90\% with an interval of 10\% as target missing rates.

The original OPPORTUNITY dataset itself has some parts of missing data due to the disconnection of wireless sensor devices \cite{chavarriaga2013opportunity}.
The missing rates are 8.0\%, 4.1\%, and 4.1\% for the training, validation, and test data, respectively.
In fact, there are a considerable number of multimodal datasets that have missing data, including Berkeley Multimodal Human Action Database \cite{ofli2013berkeley}, mice protein expression dataset \cite{higuera2015self}, and multimodal dataset for distracted driving \cite{taamneh2017multimodal}. 
These strongly support our research problem that handling missing data is crucial in the real-world applications for multimodal data integration.

Unlike the late integration, multimodal autoencoder, and EmbraceNet models, the early integration, intermediate integration, and compact multi-linear pooling models cannot naturally handle missing data.
For those models, we follow the procedure in the OPPORTUNITY challenge by filling the missing parts with their previous values.

\section{Results}
\label{sec:results}

We implement all the network models in TensorFlow \cite{abadi2016tensorflow}.
The Adam optimization method \cite{kingma2014adam} with ${\beta}_{1}=0.9$, ${\beta}_{2}=0.999$, $\hat{\epsilon}={10}^{-2}$, and a learning rate of ${10}^{-3}$ is used to train the models.
In addition, the dropout scheme \cite{srivastava2014dropout} with $p=0.5$ is applied to the fully connected and recurrent layers during the training procedures.
The batch size of each training iteration is set to 64.

\begin{figure*}[ht!]
	\centering
	\subfigure[]{
		\includegraphics[width=3.2in]{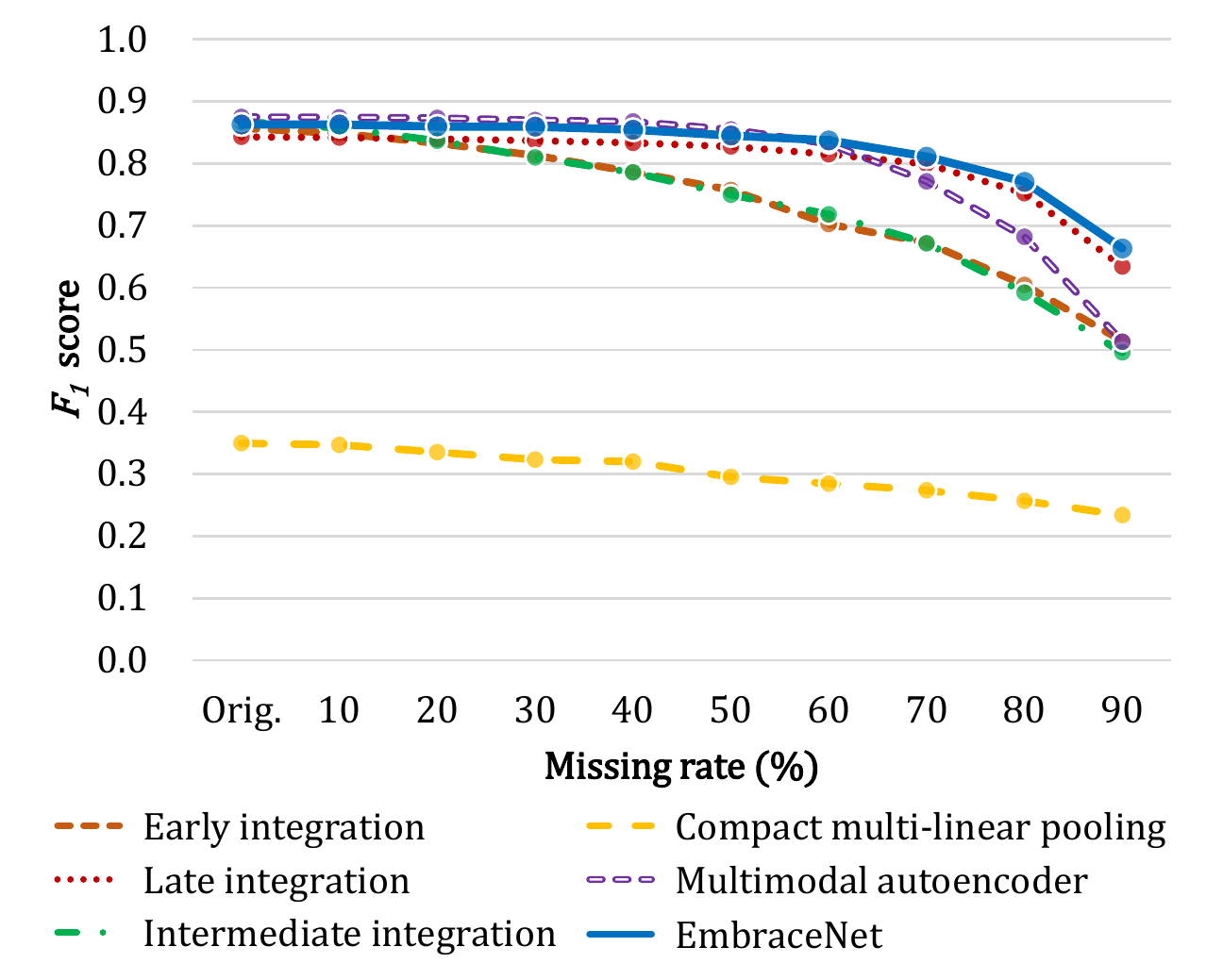}
	}
	\hspace{0.1in}
	\subfigure[]{
		\includegraphics[width=3.2in]{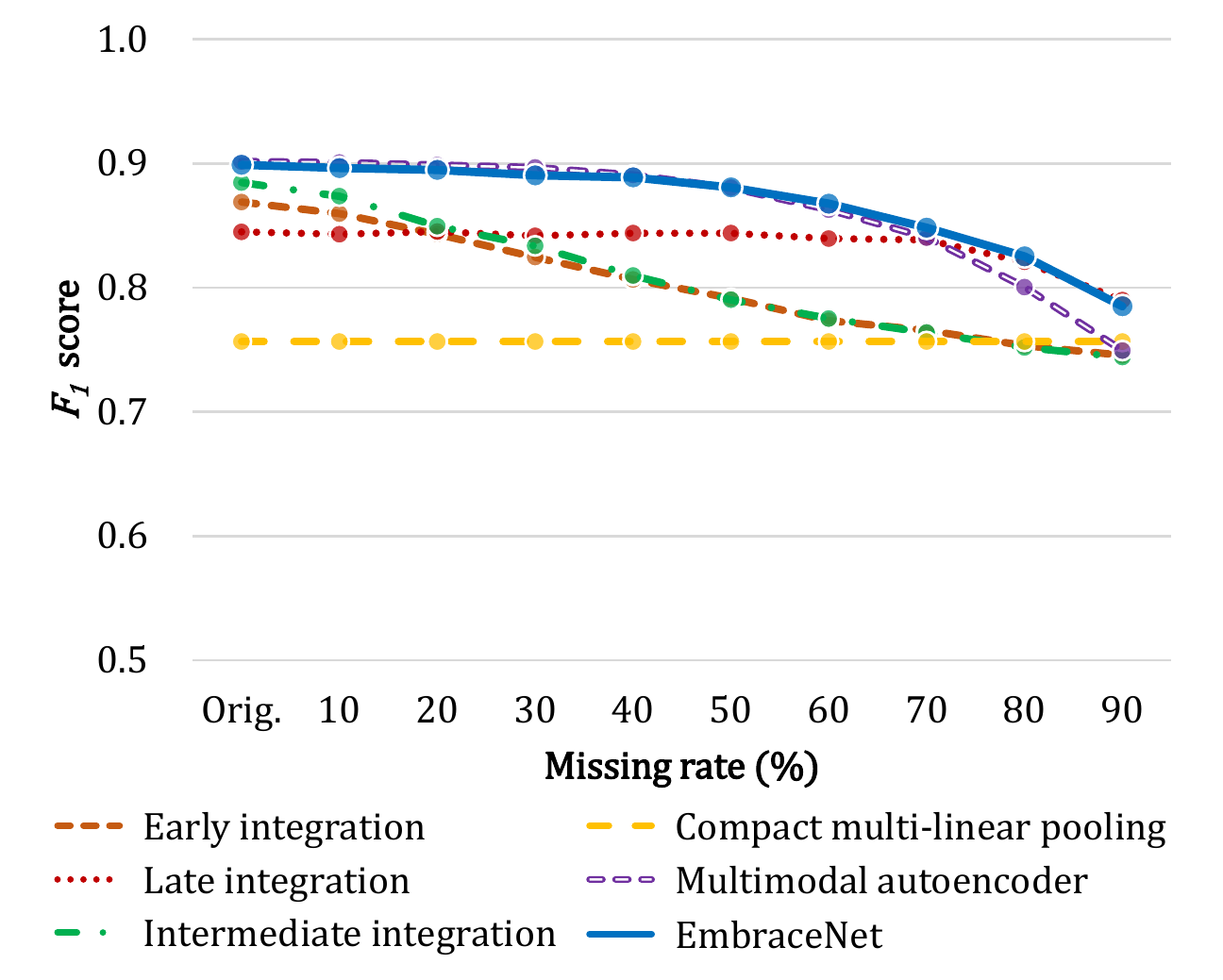}
	}
	\caption{$F_1$ scores of the six models with respect to different proportions of missing data for the classification tasks of the OPPORTUNITY dataset. (a) Modes of locomotion (b) Gestures}
	\label{fig:result_blockmissing}
\end{figure*}

\begin{table*}
\small
\renewcommand{\arraystretch}{1.2}
\centering
\begin{tabular}{c|ccc|ccc}
	\hline
	\hline
	\multicolumn{7}{c}{Modes of locomotion ($F_1$ scores)} \\
	\hline
	& \multicolumn{3}{c|}{Missing modalities} & \multicolumn{3}{c}{Missing block-wise data} \\
	\hline
	& Original & 80\% missing & Difference & Original & 80\% missing & Difference \\
	\hline
	Early integration & 0.897 & 0.287 & 68.0\% & 0.857 & 0.604 & 29.5\% \\
	Late integration & 0.885 & 0.793 & 10.4\% & 0.843 & 0.752 & 10.8\% \\
	Intermediate integration & 0.903 & 0.299 & 66.9\% & 0.868 & 0.592 & 31.7\% \\
	Compact multi-linear pooling & 0.421 & 0.247 & 41.3\% & 0.350 & 0.257 & 26.6\% \\
	Multimodal autoencoder & 0.917 & 0.696 & 24.1\% & 0.875 & 0.682 & 22.1\% \\
	EmbraceNet & 0.912 & 0.810 & 11.2\% & 0.863 & 0.771 & 10.7\% \\
	\hline
	\hline
	\multicolumn{7}{c}{Gestures ($F_1$ scores)} \\
	\hline
	& \multicolumn{3}{c|}{Missing modalities} & \multicolumn{3}{c}{Missing block-wise data} \\
	\hline
	& Original & 80\% missing & Difference & Original & 80\% missing & Difference \\
	\hline
	Early integration & 0.843 & 0.709 & 15.9\% & 0.869 & 0.753 & 13.3\% \\
	Late integration & 0.806 & 0.785 & 2.6\% & 0.845 & 0.821 & 2.9\% \\
	Intermediate integration & 0.859 & 0.709 & 17.5\% & 0.885 & 0.752 & 15.0\% \\
	Compact multi-linear pooling & 0.707 & 0.707 & 0.0\% & 0.757 & 0.757 & 0.0\% \\
	Multimodal autoencoder & 0.877 & 0.760 & 13.4\% & 0.902 & 0.800 & 11.2\% \\
	EmbraceNet & 0.873 & 0.782 & 10.4\% & 0.899 & 0.825 & 8.2\% \\
	\hline
	\hline
\end{tabular}
\caption{Comparison of classification performance of the six deep learning architectures with respect to different scenarios in the OPPORTUNITY dataset. $F_1$ scores for the original and missing data and the relative difference between them (in \%) are shown for each configuration. The proportion of the missing data is equally set to 80\%.}
\label{table:fiftymissing_opportunity}
\end{table*}

\subsection{Missing modalities}
\label{sec:results_missingmodalities}

Figures~\ref{fig:result_sensormissing_gas} and \ref{fig:result_sensormissing_opportunity} show the classification performance with respect to different numbers of modalities for the gas sensor arrays and OPPORTUNITY datasets, respectively.
Overall, the EmbraceNet model mostly outperforms the other network models.
When some of the modalities are lost, the EmbraceNet model shows much better performance than the early integration, late integration, intermediate integration, and compact multi-linear pooling models.
When only a few modalities are available, the EmbraceNet model also significantly outperforms the multimodal autoencoder model.

The early integration, intermediate integration, and compact multi-linear pooling models show much lower $F_1$ scores than the other models in most cases.
For example, in \figurename~\ref{fig:result_sensormissing_gas}, the performance of these models significantly degrades as the number of modalities decreases and plummets to $F_1$ scores of almost 0.1 when only one modality is available.
It confirms that the conventional methods without considering absence of some modalities are highly vulnerable to the loss of data.
Note that they keep performance to almost $F_1$ scores of 0.7 on classifying gestures of the OPPORTUNITY dataset (\figurename~\ref{fig:result_sensormissing_opportunity} (b)), only because they try to classify any input data to the \textit{null} class, which accounts for about 70\% of the dataset.

The compact multi-linear pooling method shows the worst performance than the other methods in all cases.
During the training steps, we observe that the loss values of the compact multi-linear pooling models occasionally increase due to the instability of the integration part: when some values of the Fourier-transformed count sketches are large, the element-wise multiplication process significantly amplifies the corresponding outputs, which is more serious when a larger number of modalities are involved.
Therefore, although the method shows good performance in the tasks with two or three modalities \cite{gao2016compact,algashaam2017multispectral,fukui2016multimodal}, it is not effective for the cases with larger numbers of modalities.

When only one modality is available, there is no correlated information between modalities to exploit.
Thus, the late integration models, which independently concentrate only on the characteristics of single modalities, show slightly better performance than the other models, which handle characteristics of both single and combined modalities.
Note that the availability of data for only one modality would be highly rare in the practical scenarios, where almost all the data are missing, e.g., about 95\% of data for the OPPORTUNITY dataset.
However, when more than two modalities are involved, the EmbraceNet models outperform the late integration models in all classification tasks, since correlated information between modalities becomes available.
The performance gap between the late integration and EmbraceNet models is more prominent in the tasks classifying the gas types (\figurename~\ref{fig:result_sensormissing_gas}) and the gestures (\figurename~\ref{fig:result_sensormissing_opportunity} (b)).

In the human activity recognition tasks, classification of the modes of locomotion is much easier than classification of the gestures.
While there are only five classes in the locomotion classification task, there are 18 classes in the gesture classification task.
In addition, more than 70\% of the data are assigned to the \textit{null} class in the latter task, which may make the network models overfit to the \textit{null} class.
Because of that, both the late integration and compact multi-linear pooling models fail to learn the characteristics of the modalities, which lead to much lower performance than the other models when multiple modalities are involved.

To further investigate the behavior of the EmbraceNet model, we examine the output values of the model.
\figurename~\ref{fig:result_embraced_representations_gas} shows averaged activations of the embracement layer, which is the final output of the EmbraceNet architecture, with respect to different classes and different numbers of modalities for the gas sensor arrays dataset.
The activation values are obtained for the whole test data, regardless of whether the classification result of each data is correct or not.
Each output vector having a length of 1024 is reshaped to a matrix having a size of $32\times32$ for better visualization.
The figure shows that the embracement layer produces consistent output values for the same target class for different numbers of modalities, even when only one or two modalities are given.
In addition, the output patterns of the layer are highly distinguishable for different classes, which confirms that the layer finely differentiates the classes.
This demonstrates that the EmbraceNet architecture robustly handles multiple modalities and prevents it from being highly dependent only on the characteristics of specific modalities.

\subsection{Missing block-wise data}
\label{sec:results_blockmissingdata}

\figurename~\ref{fig:result_blockmissing} compares the classification performance of the network models with respect to different missing rates for modes of locomotion and gestures in the OPPORTUNITY dataset.
Similarly to the results shown in Section~\ref{sec:results_missingmodalities}, the EmbraceNet models outperform the other ones in most cases.

All the network models for the classification of locomotion achieve similar performance for the original dataset without missing data, except for the compact multi-linear model.
These models yield $F_1$ scores of at least 0.84, which are similar to the results reported in the OPPORTUNITY challenge \cite{chavarriaga2013opportunity}.
This confirms that these network models have suitable architectures and are properly trained\footnote{The original DeepConvLSTM model \cite{ordonez2016deep} reported slightly better performance, but it did not precisely follow the protocol of the OPPORTUNITY challenge and thus its performance cannot be directly compared.}.

When the block-wise missing is introduced, the network models tend to show lowered performance due to the reduced information.
This tendency is more prominent in the early and intermediate integration models than the other ones, which is similar to the results in Section~\ref{sec:results_missingmodalities}.
For example, when an aggressive block-wise missing with a rate of 90\% is applied, the early and intermediate integration models record $F_1$ scores of 0.514 and 0.496 for classifying the modes of locomotion, respectively.
On the other hand, the EmbraceNet model robustly maintains the performance for all cases.
For example, when a half of the sensor data are missing (i.e., a missing rate of 50\%), the $F_1$ scores of the EmbraceNet models are only reduced by 0.017 and 0.018 for classifying the locomotion and gestures, respectively.

Table~\ref{table:fiftymissing_opportunity} shows the $F_1$ scores of the network models with respect to the original and missing data for both scenarios (i.e., missing modalities and missing block-wise data) in the OPPORTUNITY dataset.
To compare the two scenarios with the same proportion of missing data, the $F_1$ scores measured from 20\% of the modalities are used for the scenario of missing modalities, and the $F_1$ scores for a missing rate of 80\% are shown for the scenario of missing block-wise data.
The results show that the relative reductions of the $F_1$ scores for block-wise missing data are smaller than those for missing modalities, except for the late integration models.
For instance, for classifying the modes of locomotion, the $F_1$ score of the early integration model for the block-wise missing data is 29.5\% lower than that for the original data, while that for the missing modalities is 68.0\% lower.
While the classification with missing modalities completely removes information of specific modalities, removing parts of the data maintains some information of each modality, which results in partially preserving consideration of correlated characteristics.
It can be observed that, the EmbraceNet models not only show significantly smaller differences between the $F_1$ scores measured on the original and missing data, but also reveal remarkable robustness against loss of some modalities entirely.
For example, for classification of the locomotion, the early and intermediate integration models have about 2.3 and 2.1 times larger relative reductions of the $F_1$ scores obtained with missing modalities than those with block-wise missing data, respectively, whereas the EmbraceNet model achieves almost the same relative reductions.
On the other hand, the late integration models show more degraded results on missing block-wise data than missing modalities, since they do not consider any cross-modal information.
These confirm that the EmbraceNet model successfully preserves the performance against various scenarios of data loss.

\section{Conclusion}
\label{sec:conclusion}

In this paper, we proposed a novel deep learning architecture for multimodal information, called ``EmbraceNet,'' which is robust to loss of data and modalities in the wild.
Our model employs the so-called \textit{embracement} process, which not only integrates multimodal information with considering cross-modal correlations, but also robustly deals with absence of part of data.
We showed three major benefits of the EmbraceNet model, and discussed additional optimization techniques with a toy experiment on a bimodal dataset.

In addition, we compared our model with other state-of-the-art architectures, and showed the improved robustness of our model.
Details of the advantages that we showed in this paper can be summarized as follows.
\begin{itemize}
	\item
	We showed the high compatibility of the EmbraceNet model with various types of networks.
	The EmbraceNet model \textit{embraces} convolutional, dense, or recurrent neural networks by attaching our architecture between the feature extraction and classification parts.
	\item 
	We showed that our model has a capability to effectively learn cross-modal information.
	Therefore, when no data or modality is missing, it achieved similar or higher performance than the early and intermediate integration methods that take into account the correlations between modalities, while the performance of the late integration model was even lower than that.
	\item 
	We showed that the EmbraceNet model has robustness for limited availability of data.
	Thanks to the architecture designed carefully to handle multimodal data, our model kept good performance against both block-wise missing data and missing modalities.
\end{itemize}

\section*{Acknowledgements}

This research was supported by the MSIT (Ministry of Science and ICT), Korea, under the ``ICT Consilience Creative Program'' (IITP-2018-2017-0-01015) supervised by the IITP (Institute for Information \& communications Technology Promotion). In addition, this work was also supported by the IITP grant funded by the Korea government (MSIT) (R7124-16-0004, Development of Intelligent Interaction Technology Based on Context Awareness and Human Intention Understanding).


\bibliography{infofusion}

\end{document}